\documentclass{article}
\PassOptionsToPackage{numbers}{natbib}
\usepackage[preprint]{neurips_2019}
\usepackage[utf8]{inputenc} 
\usepackage[T1]{fontenc}    
\usepackage{xcolor}	
\usepackage{amsmath}
\usepackage{amsfonts}
\usepackage{nicefrac}
\usepackage{hyperref}
\usepackage{cleveref}
\usepackage{graphicx}
\usepackage{subcaption}
\usepackage{placeins}
\usepackage{booktabs}
\usepackage{algorithm}
\usepackage{algorithmicx}
\usepackage{algpseudocode}

\newcommand{\ignore}[1]{}
\usepackage[toc,page]{appendix}

\newcommand{\myvec}[1]{\boldsymbol{#1}}
\newcommand{\mymat}[1]{\boldsymbol{\mathrm{#1}}}
\newcommand{\myset}[1]{\mathcal{#1}}
\newcommand{\expectation}{\mathbb{E}}
\newcommand{\model}{CogKR}
\newcommand{\fullmodel}{Cognitive KG Reasoning}
\newcommand{\real}{\mathbb{R}}
\newcommand{\kg}{\mathcal{G}}
\newcommand{\relationset}{\mathcal{R}}
\newcommand{\entityset}{\mathcal{E}}
\newcommand{\entity}{e}
\newcommand{\headentity}{h}
\newcommand{\tailentity}{t}
\newcommand{\relation}{r}

\newcommand{\fewnum}{m}
\newcommand{\similarity}{f}
\newcommand{\queryentity}{{\hat{\headentity}}}
\newcommand{\truetail}{\hat{\tailentity}}
\newcommand{\supporthead}{{\headentity_\relation}}
\newcommand{\supporttail}{{\tailentity_\relation}}
\newcommand{\supportpair}{{(\supporthead, \supporttail)}}
\newcommand{\summary}{S}
\newcommand{\embedding}{\myvec{v}}
\newcommand{\neighbor}{\mathcal{N}}
\newcommand{\representation}{\myvec{\omega}}
\newcommand{\coggraph}{\mathrm{G}}
\newcommand{\nodeset}{V}
\newcommand{\edgeset}{E}
\newcommand{\hidden}{\mymat{X}}
\newcommand{\noderep}[1]{{\hidden[{#1}]}}
\newcommand{\frontierset}{\myset{F}}
\newcommand{\topk}{n}
\newcommand{\maxnode}{\lambda}
\newcommand{\maxneighbor}{\eta}
\newcommand{\timestep}{i}
\newcommand{\actionset}{A}
\newcommand{\action}{a}

\newcommand{\dimension}{d}
\newcommand{\parameter}{{\myvec{\theta}}}
\newcommand{\trainset}{D}

\title{Cognitive Knowledge Graph Reasoning for One-shot Relational Learning}
\author{Zhengxiao Du$^{1}$, Chang Zhou$^{2}$, Ming Ding$^{1}$, Hongxia Yang$^{2}$, Jie Tang$^{1}$\\
$^1$ Department of Computer Science and Technology, Tsinghua University\\
$^2$ DAMO Academy, Alibaba Group\\
\texttt{\{duzx16, dm18\}@mails.tsinghua.edu.cn, jietang@tsinghua.edu.cn}\\
\texttt{\{ericzhou.zc,yang.yhx\}@alibaba-inc.com}
}

\begin{document}
\maketitle

\begin{abstract}
Inferring new facts from existing knowledge graphs (KG) with explainable reasoning processes is a significant problem and has received much attention recently. However, few studies have focused on relation types unseen in the original KG, given only one or a few instances for training. To bridge this gap, we propose \model~for one-shot KG reasoning. The one-shot relational learning problem is tackled through two modules: the summary module summarizes the underlying relationship of the given instances, based on which the reasoning module infers the correct answers. Motivated by the dual process theory in cognitive science, in the reasoning module, a cognitive graph is built by iteratively coordinating retrieval (System 1, collecting relevant evidence intuitively) and reasoning (System 2, conducting relational reasoning over collected information). The structural information offered by the cognitive graph enables our model to aggregate pieces of evidence from multiple reasoning paths and explain the reasoning process graphically. Experiments show that \model\space substantially outperforms previous state-of-the-art models on one-shot KG reasoning benchmarks, with relative improvements of 24.3\%-29.7\% on MRR\footnote{The source code is available at \url{https://github.com/THUDM/CogKR}}.
\end{abstract}

\section{Introduction}
\label{sec:introduction}
A typical knowledge graph (KG) is far from completion due to the arbitrary complex relations, making it essential to enhance the ability to infer new facts from the existing relations \cite{TransE, Complex:TrouillonDGWRB17,DeepPath:XiongHW17,VariationalReasoning:ChenXYW18} for downstream tasks, e.g., question answering \cite{QAOverKG:MohammedSL18}, dialogue system \cite{DialogueSystemYoungCCZBH18}, and relation extraction \cite{RelationExtraction:WangZWZCZZC18}.   Most previous studies focus on completing facts of existing relation types in KG. However, as the existing relation types in a KG are always limited, the ability to infer the facts of an unseen relation is critical but not well studied. Large amounts of training instances of that relation are required for traditional KG completion approaches. 

The remedy to discover facts for new relations in a data-efficient way could be related to one-shot learning, which has been proposed in the image classification task \cite{OneShot:Fei-FeiFP06}, trying to recognize objects in a new class given only one or a few instances of that class.
Similarly, one-shot relational learning for KG, which aims to uncover facts of a new relation given only one training instance, has been proposed recently by \citet{Oneshot-Relational:XiongYCGW18}. 
Nevertheless, one-shot learning relies heavily on strong prior knowledge learned from previous classes, as the training set for a single class becomes minimal. The form of prior knowledge represents the inductive bias of the learning algorithm.
The success achieved for one-shot learning in image data stems from the design of multi-level feature extraction architectures inspired by the human visual perception system. 
On the contrary, KG reasoning, whose underlying data is relational and discrete, is known to be related to the cognitive system \cite{battaglia2018relational}, since human cognition also works with strong prior knowledge that our world consists of objects and relations \cite{spelke2007core}. 

A proper inductive bias of inferring facts for new relations, we believe, could be found in the cognitive system of human beings.
As suggested in the dual process theory \cite{theempirical:Sloman96}, the reasoning system of human beings consists of two distinct processes, one to retrieve relevant information via an implicit and unconscious system (System 1) and the other to reason over the collected information via an explicit, conscious and controllable reasoning process (System 2). Compared with the case for image perception, training a relational inference system that learns the underlying reasoning process of human beings could be a better choice for learning new relations. 
The method by \citet{Oneshot-Relational:XiongYCGW18} can be viewed as a System 1 only approach, which learns an implicit matching metric between entity pairs based on KG embeddings \cite{TransE}.
It focuses too much on the similarity, rather than the relationship, leading to incorrect facts. Similar conclusions can be found in reading comprehension tasks that require reasoning over the discrete input data \cite{jia2017adversarial}. This misjudgment ratio will inevitably rise as the reasoning complexity increases due to the lack of System 2.
System 2 requires explicit reasoning capacity, which has been studied in the field of KG reasoning methods \cite{VariationalReasoning:ChenXYW18,DeepPath:XiongHW17,GoForAWalk,RewardShaping:LinSX18}. Most methods use random walk \cite{PRA:LaoMC11,ChainOfReasoning:McCallumNDB17} or path-finding policy learned by RL \cite{DeepPath:XiongHW17,GoForAWalk} to obtain paths connecting two entities and infer the relationship from the paths with neural networks.
However, despite that they are not studied in one-shot relational learning, the expressiveness of a single path is limited, compared to a subgraph which keeps the clues complete to reason.

In this paper, we propose \fullmodel~(\model), in which a summary module and a reasoning module work together to address the one-shot KG reasoning problem. 
In the summary module, the knowledge about the entities in the training instance is collected, and their underlying relation is represented as a continuous vector. Then in the reasoning module, new facts of the relation are inferred based on the relation vector and the KG. 
Specifically, we design a new method to reason out the missing tail entity of a relation given the head entity, under the inspiration of the dual process theory \cite{theempirical:Sloman96} in cognitive science. At each step, query-relevant entities and relations are retrieved from the neighborhood and organized as a cognitive graph \cite{ding2019cognitive}, which resembles the capacity-limited working memory \cite{baddeley1992working}. Then relational reasoning is conducted over the graph to update the nodes' representations. The above process is iterated until all relevant evidence is found. Then the final answer is predicted based on the reasoning results. 
\section{Problem Formulation}
\label{sec:preliminary}

A knowledge graph $\kg$ is represented as a set of triples $\{(\headentity, \relation, \tailentity)\}\subseteq\entityset\times\relationset\times\entityset$. Each triple consists of a relation $\relation\in\relationset$ and two entities $\headentity,\tailentity\in\entityset$, which denotes a directed edge of type $\relation$ from $\headentity$ to $\tailentity$. On the KG, the one-shot KG reasoning problem can be formalized as: given a few entity pairs $\{(\headentity_{\hat{r}}^{(k)}, \tailentity_{\hat{r}}^{(k)})\}_{k=1}^\fewnum$ of an unseen relation type $\hat{r}$, we would like to predict the tail entity $\hat{\tailentity}$ of a missing triple $(\queryentity, \hat{r}, ?)$. In this paper, we mainly focus on the case when only one pair $(\headentity_{\hat{r}}, \tailentity_{\hat{r}})$ is given, i.e. $\fewnum=1$. However, our method can also be extended to few-shot cases by existing few-shot learning methods \cite{MatchingNetwork:Vinyals, ProtoNet:Snell}. We define the probability $p_\parameter(\cdot|\queryentity,(\supporthead, \supporttail))$ over the entity set $\entityset$ as the probability of entities to be the correct answer given the support pair $(\supporthead, \supporttail)$ of relation $\relation$ and query head entity $\queryentity$. ${\boldsymbol{\parameter}}$ are the parameters of our model and represent the apriori learned from the existing facts. The training objective should maximize $p_\parameter(\truetail|\queryentity, (\supporthead, \supporttail))$ given the ground truth $\truetail$:
\begin{equation}
\label{eqn:objective}
\max_{\parameter} \expectation_{\relation\in\relationset}\left[\expectation_{(\queryentity, \truetail)\in\trainset_\relation,(\supporthead,\supporttail)\in\trainset_\relation}\left[{p_\parameter(\truetail|\queryentity,(\supporthead,\supporttail))} \right]\right]
\end{equation}
where $\trainset_\relation=\{(\headentity,\tailentity)|(\headentity,\relation,\tailentity)\in\kg\}$ is the set of entity pairs for relation $r$.

\section{Approach}
\label{sec:approach}
\begin{figure}[!ht]
\centering
\includegraphics[width=0.92\textwidth]{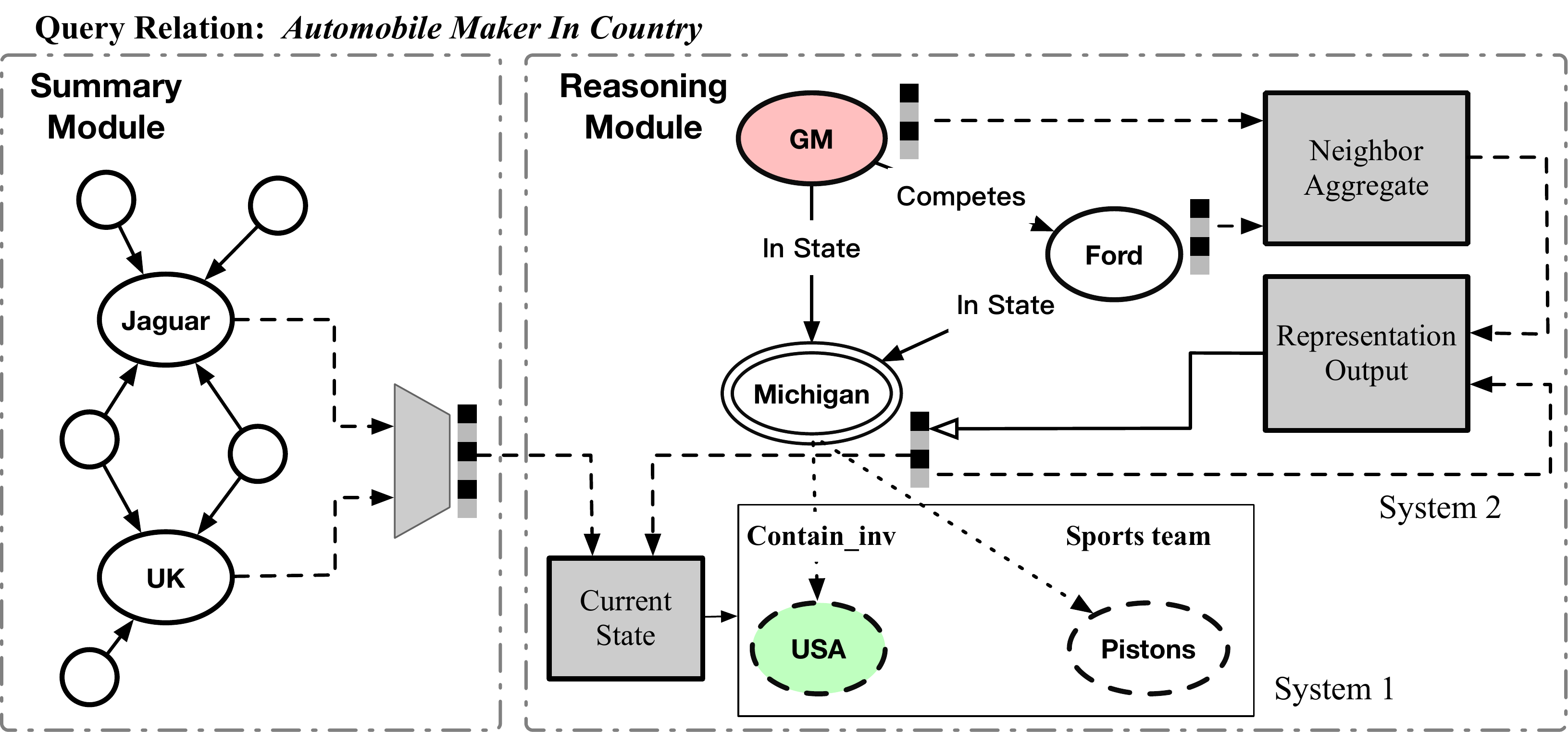}
\caption{Overview of the \model\space framework. The summary module generates the vector for the underlying relationship between the support pair (Jaguar, UK). The reasoning module builds the cognitive graph on the right. When updating the hidden representation of Michigan in System 2, representations of its ingoing edges are aggregated and combined with its entity embedding. Then in System 1, the outgoing edge Contains\_inv of Michagan, which leads to the correct answer USA, is added to the graph.}
\label{fig:framework}
\end{figure}
In this section, we describe the proposed model for the one-shot KG reasoning, the training algorithm, and the complexity analysis. Our framework for the one-shot problem consists of two modules. The first module is called \emph{Summary Module}, which maps an entity pair $\supportpair$ to a continuous representation $\representation_{\supporthead, \supporttail}$ of their underlying relation. Given only one training instance for the query relation, the mapping learned by the neural network can generalize better than direct optimization. The second stage is called \emph{Reasoning Module}, which, given the representation $\representation_{\supporthead, \supporttail}$ and a head entity $\queryentity$, predicts the correct tail entity $\truetail$. Similar to the reasoning process of humans, the reasoning module combines implicit retrieval and explicit reasoning, by expanding and reasoning over the cognitive graph iteratively. The overview of the whole framework is shown in \Cref{fig:framework}.

\subsection{Summary Module}
In the summary module, we collect information about the entities in the given instance and summarize the relationship between them. Previous works \cite{DeepPath:XiongHW17, GoForAWalk} demonstrated that the relationship of two entities could be inferred from paths connecting them. However, the number of paths connecting two entities can increase exponentially as the path length increases. With only one pair given, the search space cannot be effectively reduced with the prior knowledge \cite{DeepPath:XiongHW17}. 
Therefore, we apply a neural network to infer the entity pair's relationship from their vector representations, which are generated by a graph neural network (GNN). Given an entity $\entity$, GNN generates the entity vector $\myvec{\omega}_\entity$ from the entity's embedding and its neighbor set:
\begin{align}
	\myvec{\omega}_e &= \sigma(\mymat{W}_{s}\embedding_e+\myvec{b}_s+\mymat{W}_c\cdot\frac{1}{|\neighbor_e|}\sum_{(\relation_k,\entity_k)\in\neighbor_e}[\embedding_\relation,\embedding_\entity]),
\end{align}
where $\mymat{W}_s, \mymat{W}_c\in\real^{d\times d}, \myvec{b}_s\in\real^{d}$ are parameters and $\neighbor_\entity=\{(\relation_k,\entity_k)|(\entity,\relation_k,\entity_k)\in\kg\}$ is the set of outgoing edges of entity $\entity$ in $\kg$. The entity and relation embeddings $\embedding_\entity$ and $\embedding_\relation$ can be pre-trained with existing KG embedding methods. Given the vector representations of an entity pair $\supportpair$, $\representation_\supporthead$ and $\representation_\supporttail$, we can get the representation of their underlying relation as $\myvec{\omega}_{\supporthead, \supporttail} = \sigma(\mymat{W}_o[\myvec{\omega}_\supporthead, \myvec{\omega}_\supporttail]+\myvec{b}_o)$, where $\mymat{W}_o\in\real^{2d\times d}, \myvec{b}_o\in\real^d$ are parameters.
\subsection{Reasoning Module}
\textbf{Cognitive Graph}\hspace{2mm} Inspired by the reasoning process of humans, the reasoning module consists of two iterative processes, retrieving information from KG (System 1) and reasoning over collected information (System 2). We use a unique structure, called \emph{cognitive graph}, to store both retrieved information and reasoning results. A cognitive graph is a subgraph of $\kg$, with hidden representations for its nodes. The hidden representations stand for the understandings of all the entities in $\coggraph$. Formally, $\coggraph=(\nodeset,\edgeset,\hidden)$, where $\nodeset\subseteq\entityset$, $\edgeset\subseteq\kg$, and $\hidden\in\real^{|\nodeset|\times\dimension}$ is the matrix of hidden representations. The hidden representation of node $\entity$ is denoted as $\noderep{\entity}$. In the beginning, $\coggraph$ only contains the query entity $\queryentity$, which is marked as unexplored. The advantages of the cognitive graph against previous path-based reasoning methods are two-fold. On the one hand, the graph structure allows more flexible information flow.\ignore{ For example, pieces of information from different paths can be aggregated for intermediate nodes.} On the other hand, the search for correct answers can be more efficient when organized as a graph instead of paths. 

\textbf{System 1}\hspace{2mm} To retrieve relevant evidence from $\kg$, at each step $\timestep$, we select an unexplored node $\entity_\timestep$ from the subgraph, expand $\coggraph$ with part of $\entity_\timestep$'s outgoing edges, and mark $\entity_\timestep$ as explored. Given the current entity $\entity_\timestep$, the set of possible actions $\actionset_\timestep$ consists of the outgoing edges of $\entity_\timestep$ in $\kg$. Concretely, $\actionset_\timestep=\{(\relation,\entity)|(\entity_\timestep,\relation,\entity)\in\kg\}$. Following \cite{GoForAWalk}, we augment $\kg$ with the reversed links $(\tailentity, \relation^{-1}, \headentity)$ and cut the maximum number of outgoing edges of an entity by a threshold $\maxneighbor$. To give the agent the option of not expanding from $\entity_\timestep$, we add a particular action that represents no action. The actions $\myvec{\action}_\timestep$ are sampled from a multinomial distribution $\operatorname{Mult}(\topk,\myvec{p}_\timestep)$. $\topk$ is a hyperparameter that represents the action budget, and $\myvec{p}_\timestep$ is the probability over $\actionset_\timestep$, predicted as:.
\begin{equation}
\label{eqn:mult_probability}
	\myvec{p}_\timestep =\operatorname{softmax}\big(\sigma(\mymat{A}'_\timestep\mymat{W}_1)\cdot\sigma(\mymat{W}_2[\noderep{\entity_\timestep},\myvec{\omega}_{\headentity, \tailentity}])\big)
\end{equation}
where $\mymat{W}_1\in\real^{3\dimension\times\dimension}, \mymat{W}_2\in\real^{\dimension\times 2\dimension}$ are parameters and $\mymat{\actionset}_\timestep\in\real^{|\actionset_\timestep|\times 3\dimension}$ is the candidate matrix which stacks the concatenated embeddings of all actions in $\actionset_\timestep$. The embedding of an outgoing edge $(\relation, \entity)$ is the concatenation of the entity's embedding $\embedding_{\entity}$, the relation's embedding $\embedding_{\relation}$, and the entity's hidden representation $\noderep{\entity}$ (filled with $0$ if $\entity$ does not belong to $\coggraph$). Notably, the embedding of "no action" is a trainable vector of length $3\dimension$.  Edges selected in $\myvec{\action}_\timestep$ are added to $\edgeset$. Nodes that are related to selected edges but not belong to $\coggraph$ are added to $\nodeset$ and marked as unexplored. To limit the size of $\coggraph$, after $|\nodeset|$ reaches the predefined maximum node number $\maxnode$, we will not add new nodes.

\textbf{System 2}\hspace{2mm} In this paper, we apply deep learning to conduct relational reasoning, which has shown better generalization capacity than rule-based reasoning for KG \cite{ChainOfReasoning:McCallumNDB17,VariationalReasoning:ChenXYW18}. After each graph expansion step, the hidden representations of related nodes are updated based on their neighbors in $\coggraph$. For a node $\entity$, the updating formula of hidden representation $\noderep{\entity}$ is:
\begin{align}
\label{eqn:neighbor_aggregate}
	 \Delta_e &= \frac{1}{|E_\entity|}\sum_{(\relation_k,\entity_k)\in E_e}\mymat{W}_3[\embedding_{\relation_k},\noderep{\entity_k}]\\
\label{eqn:hidden_update}
	\noderep{\entity} &=\sigma(\mymat{W}_4\embedding_e+\Delta_e+\myvec{b}_4)
\end{align}
where $\mymat{W}_3\in\real^{\dimension\times 2\dimension},\mymat{W}_4\in\real^{\dimension\times\dimension},\myvec{b}_4\in\real^\dimension$ are parameters and $E_\entity=\{(\relation',\entity')|(\entity',\relation',\entity)\in \edgeset \}$ is the set of ingoing edges for entity $\entity$ in $\coggraph$. This formula can be considered as a variant of GNN \cite{battaglia2018relational}: the representation of a node is computed as a combination of its own information and the aggregation of its neighbors' information. However, unlike traditional GNNs, where the current layer of representations is computed from the previous layer(s), all the representations are in the same layer but computed sequentially. It can also be considered as an extension to Path-RNN \cite{CompositionalReasoning:NeelakantanRM15}, augmented with the ability to aggregate the information from multiple paths for intermediate nodes.

\begin{algorithm}[tb]
\caption{One-shot Multi-hop KG Reasoning Algorithm}
\label{alg:reasoning}
\begin{algorithmic}[1]
\Require Entity pair $\supportpair$; Query entity $\queryentity$; KG $\kg$
\State $\representation_{\supporthead, \supporttail}\gets \summary(\supporthead, \supporttail)$ 
\State $\nodeset\gets\{\queryentity\}$, $\edgeset\gets\emptyset$, $\frontierset\gets\{\queryentity\}$
\State $\noderep{\queryentity}\gets\sigma(\mymat{W}_4\embedding_\queryentity+\myvec{b}_4)$
\Repeat
\State Pop an entity $e$ from $\frontierset$
\State Build the action set $\actionset$ for $\entity$
\State Sample actions $\myvec{\action}$ from multinomial distribution over $\actionset$ 
\For{$\relation',\entity'$ in $\myvec{\action}$}
\State $\edgeset\gets\edgeset\cup\{(e, r',e')\}$
\If{$\entity'\notin\nodeset$ and $|\nodeset|<\maxnode$}
\State $\nodeset\gets\nodeset\cup\{e'\}$, $\frontierset\gets\frontierset\cup\{e'\}$
\EndIf
\State Update hidden representations $\hidden[e']$ with \Cref{eqn:neighbor_aggregate,eqn:hidden_update}
\EndFor
\Until{$\frontierset=\emptyset$}
\State \textbf{return} $\operatorname{argmax}_{e\in\nodeset}{\similarity(\noderep{\entity}, \representation_{\supporthead,\supporttail})}$
\end{algorithmic}
\end{algorithm}

After all the nodes in $\coggraph$ are marked as explored, we terminate the reasoning process and predict nodes' probability to be the correct answer based on their hidden representations:
\begin{align}
	\similarity(\noderep{\entity}, \representation_{\supporthead, \supporttail} ) &= \mymat{W}_p[\noderep{\entity},\representation_{\supporthead,\supporttail}]\\
	q(\entity|\coggraph, (\supporthead, \supporttail)) &= \frac{\exp(\similarity(\noderep{\entity}, \representation_{\supporthead, \supporttail}))}{\sum\limits_{\entity'\in\coggraph}\exp(\similarity(\noderep{\entity'}, \representation_{\supporthead, \supporttail}))}, \entity\in\coggraph
\end{align}
where $\mymat{W}_p\in\real^{\dimension\times 2\dimension}$ are parameters. The complete algorithm is presented in \Cref{alg:reasoning}. In the algorithm, a queue $\frontierset$ is used to store the unexplored nodes.

\subsection{Optimization}
Based on previous subsections, we can write the probability in \Cref{eqn:objective} as\footnote{We leave out $\parameter$ in the subscripts for simplicity.} :
\begin{equation}
	\label{eqn:probability}
	p(\truetail|\queryentity, (\supporthead, \supporttail))=\expectation_{\coggraph\sim\pi(\queryentity, (\supporthead, \supporttail))}\left[\mathbb{I}(\truetail\in\coggraph) q(\truetail|\coggraph, (\supporthead, \supporttail))\right]
\end{equation}
where $\pi(\queryentity, \supportpair)$ is the stochastic policy to build $\coggraph$ in the previous subsection. We divide the optimization of this probability into two parts, to optimize $\pi(\queryentity, \supportpair)$ and $q(\truetail|\coggraph, \supportpair)$ separately. Directly optimizing $\pi(\queryentity, \supportpair)$ requires back-propagating through random samples, which is intractable. Instead, we model the graph building with reinforcement learning. The terminal reward is $r(\coggraph)=\mathbb{I}(\truetail\in\coggraph) q(\truetail|\coggraph, (\supporthead, \supporttail))$. The latter part in $r(\coggraph)$ is dependent on $\parameter$, and in practice, we found it causes severe instability during training, so we finally leave this term out by setting $r(\coggraph)=\mathbb{I}(\truetail\in\coggraph)$. We employ the REINFORCE algorithm\cite{REINFORCE:Williams92}, to update $\parameter$ with the stochastic gradient $\nabla\parameter_{graph}=r(\coggraph)\nabla_\parameter\log\pi_\parameter(\coggraph)$. To optimize $q(\truetail|\coggraph, \supportpair)$ is to maximize the predicted probability of the correct answer in $\coggraph$. We employ the cross-entropy loss and update $\parameter$ with as $\nabla\parameter_{predict}=\mathbb{I}(\truetail\in\coggraph)\nabla_\parameter{\log{q(\truetail|\coggraph, \supportpair)}}$. During training, the gradient is added together, and we use stochastic gradient descent to approximate the gradient descent on the full dataset.

\subsection{Complexity Analysis}  
To complete a query $(\queryentity, \relation, ?)$, embedding-based methods need to enumerate the whole entity set, so it takes $O(|\entityset|)$ time for every query. For a large KG containing millions of entities combined with complex score functions, this can be highly computationally expensive. \model, on the other hand, utilizes the local structure of KG to reduce the time complexity. As for System 1, each node in $\coggraph$ is explored only once. The graph expansion step is conducted $O(|\nodeset|)$ times and each time we compute scores for at most $\maxneighbor$ outgoing edges. Therefore it takes $O(\maxneighbor|\nodeset|)$ time to complete graph expansion. Similarly the representation update in System 2 takes at most $O(\eta|\edgeset|)$ time. Finally, for prediction, we compute scores for nodes in $\nodeset$, which takes $O(|\nodeset|)$ time. As we have $|E|=O(|V|^2)$ and in practice $|\nodeset|\le\maxnode$, which is a predefined constant, \model\space takes the constant time that does not depend on the entity number, and can more easily scale up to large KGs.
\section{Experiment}
\label{sec:experiment}
\subsection{Experiment Setting}

\textbf{Datasets}\hspace{2mm} We use the NELL-One and Wiki-One datasets released by \cite{Oneshot-Relational:XiongYCGW18} for one-shot relational learning for evaluation. Both datasets are based on real-world KGs (NELL \cite{NELL-aaai15} and Wikidata \cite{Wikidata:VrandecicK14}) and created with a similar process: relations with less than 500 but more than 50 triples are selected as one-shot tasks and the background KGs are built with facts of other relations. The dataset statistics are shown in the appendix. Note that the Wiki-One dataset is an order of magnitude larger than any other benchmark datasets in terms of the number of entities and relations. In practice, we found that the Wiki-One dataset suffers from sparsity and non-connectivity in the backend KG. To better evaluate the reasoning ability, we remove 41.3\% evaluation facts whose entity pairs' distances are no less than 5 in Wiki-One. We also analyze the influence of entity pairs' distances in \Cref{subsec:quantitative}. The reason for not using standard benchmarks for KG completion, such as FB15k-237 \cite{FB15k:ToutanovaCPPCG15} and WN18RR \cite{ConvE:DettmersMS018} is that these datasets are subsets of the real-world KGs and do not contain enough relation types to train and evaluate one-shot learning algorithms.

\textbf{Baselines}\hspace{2mm} We compare \model\space with various state-of-the-art models using HITS@1,5,10 and mean reciprocal rank(MRR), which are standard metrics for KB completion tasks. For embedding based models, we compare with TransE \cite{TransE}, DistMult \cite{DistMult:Yang2015}, ComplEx \cite{Complex:TrouillonDGWRB17}, ConvE \cite{ConvE:DettmersMS018}, and TuckER \cite{TuckER}. For reasoning based models, we compare with MultiHopKG \cite{MultihopKG:LinSX18}, which outperforms MINERVA \cite{GoForAWalk}. We also compare with GMatching \cite{Oneshot-Relational:XiongYCGW18}, which is designed for one-shot KG completion and achieves impressive improvements over embedding based methods.

More details about the experiment settings can be found in the appendix.

\subsection{Performance}
\label{subsec:performance}
\Cref{tab:nell} reports the one-shot KG completion performance on NELL-One and Wiki-One datasets. Considering the relatively small scale of NELL-One, we run each method three times and report the mean and the standard deviation. ConvE, TuckER, and MultihopKG did not scale to the Wiki-One, which contains millions of entities, so their results on Wiki-One are not included.

On both datasets, \model\space outperforms previous works in all selected metrics. The improvements are particularly substantial in terms of Hits@1 and MRR. The improvements are 5.6\% and 5.0\% on NELL-One, and 7.9\% and 6.6\% on Wiki-One. We also note that GMatching, although designed for one-shot relational learning, cannot perform better than embedding-based methods on the small dataset. With similar one-shot learning settings, our method can beat their method with large margins. Compared with the reasoning-based method, our method also achieves relative improvements of 23.8\% on MRR. On the large dataset Wiki-One, we find that embedding-based methods cannot work well. Their performance is far below those of GMatching and \model.

\begin{table}
	\caption{One-shot KG reasoning results for NELL and Wikidata.}
	\label{tab:nell}
	\centering
	\resizebox{\linewidth}{!}{
	\begin{tabular}{p{21mm}| *{4}{r} | *{4}{r}}
		\toprule
		\multicolumn{1}{c|}{} & 	
		\multicolumn{4}{c|}{\textbf{NELL-One}} &
		\multicolumn{4}{c}{\textbf{Wiki-One}}\\
		\cline{2-9}\\
		Model & H@1 & H@5 & H@10 & MRR & H@1 & H@5 & H@10 & MRR\\
		\midrule
		TransE & 4.4 (0.1) & 14.9 (1.1) & 29.6 (0.5) & 11.1 (2.5) & 2.5 & 4.3 & 5.2 & 3.5 \\
		ComplEx & 9.4 (0.6) & 19.4 (0.3) & 23.9 (1.4) & 14.1 (0.6) & 4.0 & 9.2 & 12.1 & 6.9  \\
		DistMult & 12.3 (0.8) & 23.1 (2.6) & 26.9 (2.9) & 16.3 (1.6) & 1.9 & 7.0 & 10.1 & 4.8\\
		ConvE & 10.5 (2.4) & 23.0 (4.7) & 30.6 (4.6) & 17.0 (2.7) & --- & --- & --- & ---\\
		TuckER & 14.2 (0.6) & 22.5 (0.2) & 29.5 (0.6) & 19.4 (0.6) & --- & --- & --- & ---\\
		\midrule
		MultiHopKG & 14.9 (1.7) & 27.0 (3.9) & 31.2 (3.8) & 20.6 (2.4) & --- & --- & --- & ---\\
		\midrule
		GMatching & 13.3 (0.9) & 22.6 (1.4) & 29.6 (1.5) & 18.3 (1.0) & 17.0 & 26.9 & 33.6 & 22.2\\
		\midrule
		\model-onlyR & 18.9 (0.1) & 27.1 (0.4) & 29.8 (0.7) & 22.7 (0.4) & 18.5 & 21.5 & 23.3 & 20.0\\
		\model & \textbf{20.5} (0.5) & \textbf{31.4} (1.1) & \textbf{35.3} (0.9) & \textbf{25.6} (0.4) & \textbf{24.9} & \textbf{33.4} & \textbf{36.6} & \textbf{28.8}\\
		\bottomrule
	\end{tabular}
	}
\end{table}

\subsection{Quantitative analysis}
\label{subsec:quantitative}
\textbf{Ablation Study}\hspace{2mm} We conduct an ablation study to analyze the contributions of different components in \model. To understand the contributions of the summary module and the novel reasoning method with the cognitive graph, we create a baseline, \model-onlyR, which uses the same reasoning module as \model\space but without the summary module. We can see that compared with MultihopKG, which uses a path-based reasoning method, \model-onlyR can perform better on Hits@1 and MRR while achieving comparable results on Hits@5 and Hits@10, which proves the superior reasoning capacity of the proposed reasoning method. The complete model \model\space can outperform the reasoning-only module, showing the contribution of the summary module. 

\begin{figure}
	\begin{subfigure}[t]{0.49\linewidth}
		\centering
		\includegraphics[width=\linewidth]{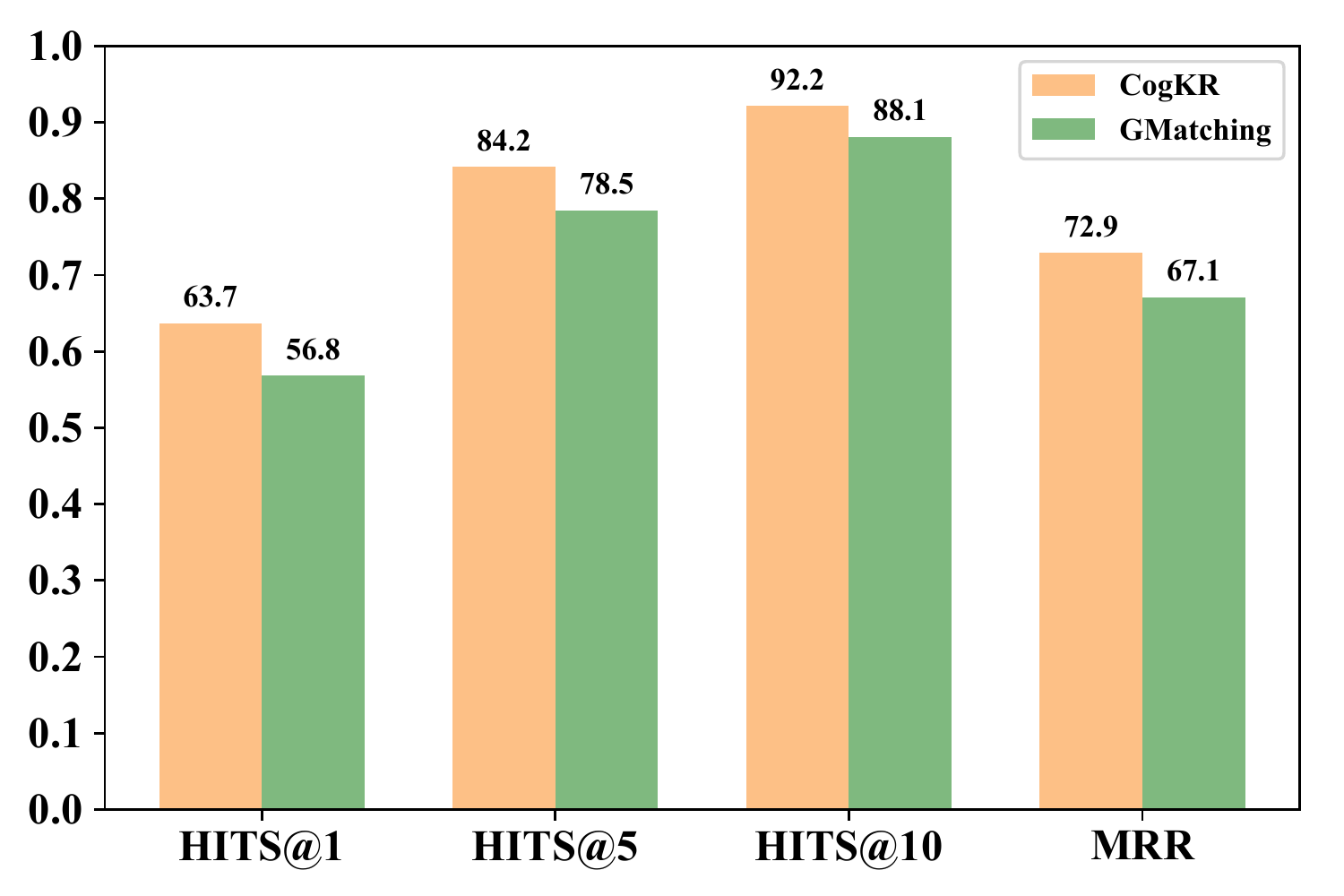}
		\caption{Scores for ranking entities in cognitive graphs.}
		\label{subfig:candidate_error}
	\end{subfigure}
	\hfill
	\begin{subfigure}[t]{0.49\linewidth}
		\centering
		\includegraphics[width=\linewidth]{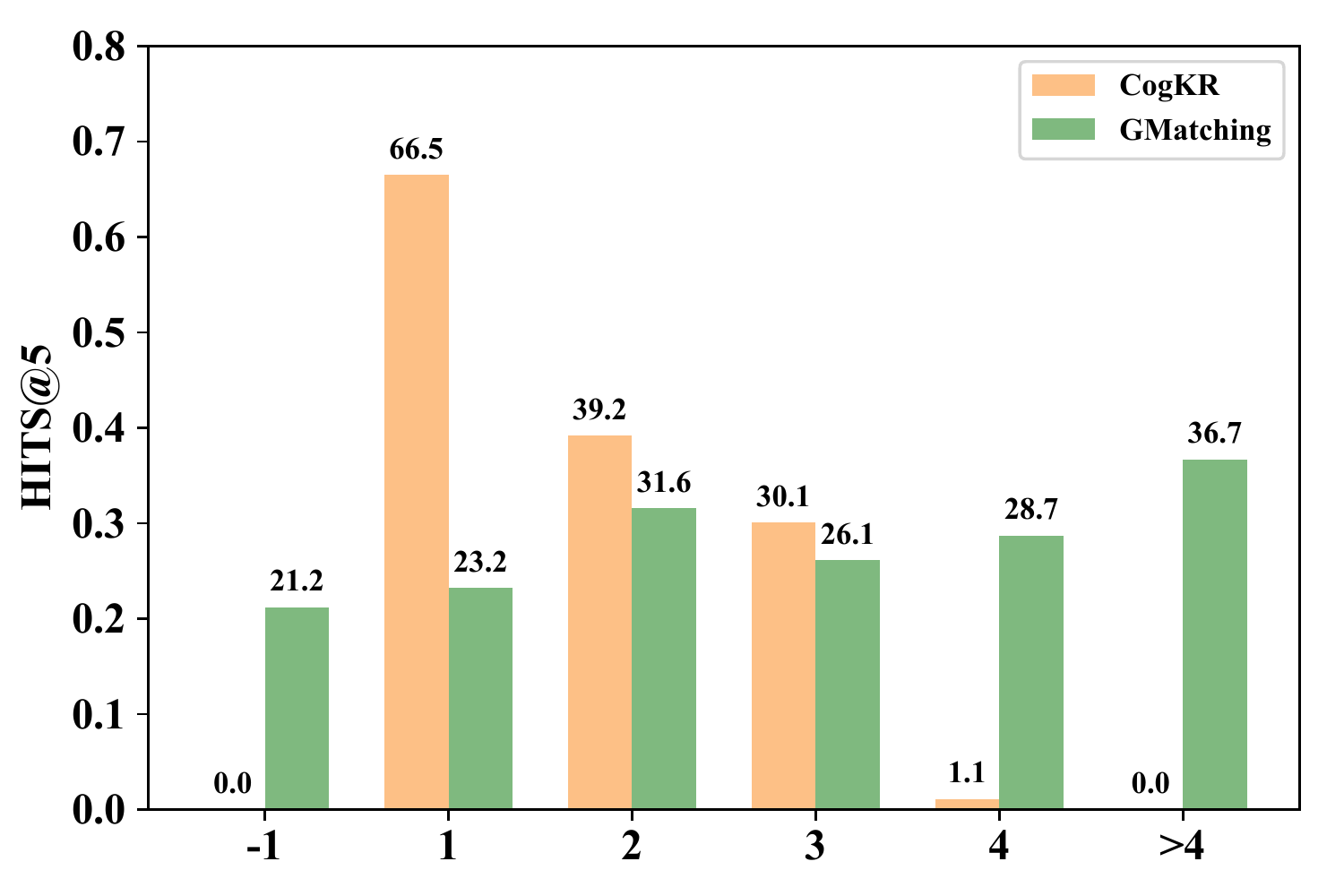}	
		\caption{HITS@5 for different shortest path lengths.}
		\label{subfig:path_length}
	\end{subfigure}
	\caption{Performance analysis of \model\space on Wiki-One.}
\end{figure}

\textbf{Strengths and Weaknesses}\hspace{2mm}
We analyze the strengths and weaknesses of \model\space compared with traditional methods on Wiki-One. We note that System 2 in \model\space only gives scores for entities that are found by System 1. To validate that System 2 based on the cognitive graph provides better reasoning ability than embedding-based methods, we compare the performance scores for ranking entities in the built cognitive graphs against GMatching in \Cref{subfig:candidate_error}. We observe that \model\space can make a more accurate prediction for entities in cognitive graphs than GMatching. Comparing the scores with those in \Cref{tab:nell}, however, we can also find 60.9\% error of HITS@1 comes from the cases that the correct answer is not found by System 1. The observation shows that a significant direction for improving \model\space is to increase the retrieval ability of System 1. In \Cref{subfig:path_length}, we show the HITS@5 against GMatching, categorized by the shortest path lengths from the query entity to the correct answer. We observe that \model\space outperforms the baseline significantly in samples whose shortest path lengths are 1, 2, or 3 steps. Given only one training instance, longer reasoning chains will be highly uncertain. However, GMatching can perform quite well in finding answers that are more than four steps away or even do not have paths at all. The reason might be they match the query entity and candidates with GNN, which can reduce the candidate space by entities' local patterns (e.x, shared relation edges for entities in training pairs and test pairs). However, from the point of logic, there is not sufficient evidence to reason the relationship of entities that are not connected in KG. If we do care about such entity pairs, the remedy is simple that we can ensemble an embedding-based method to solve the unreachable cases.

\begin{table}
	\centering
	\caption{Inference time of \model\space and GMatching for different candidate numbers.}
	\label{tab:running_time}
	\begin{tabular}{c|c|c|c|c}
		\toprule
		Candidate & \multicolumn{2}{c|}{truncated (5,000)} & \multicolumn{2}{c}{full (4,838,244)} \\
		\midrule
		Model & \model & GMatching & \model & GMatching \\
		\midrule
		Time (sec / sample) & $5.7\times 10^{-2}$ & $4.6\times 10^{-2}$ & $5.7\times 10^{-2}$ & $4.2\times 10$ \\
		\bottomrule
	\end{tabular}
\end{table}

\textbf{Running Time}\hspace{2mm} To validate \model's advantage on time complexity over baselines, we compare the inference time of \model\space and GMatching\footnote{Both models are implemented in PyTorch and tested on a single RTX 2080.}. We report the running time with truncated candidate sets and full entity sets in \Cref{tab:running_time}. We can see that when the candidate number is limited to 5000, the running time of GMatching is comparable to that of \model. However, when the candidate number is not limited, the running time of GMatching increases proportionally with the number of candidates, while the running time of \model\space remains the same. We make more discussion about the time complexity in the appendix.

\subsection{Qualitative analysis}
\label{subsec:qualitative}
\begin{figure}
	\centering
	\begin{subfigure}[t]{0.22\linewidth}
		\centering
		\includegraphics[width=\linewidth]{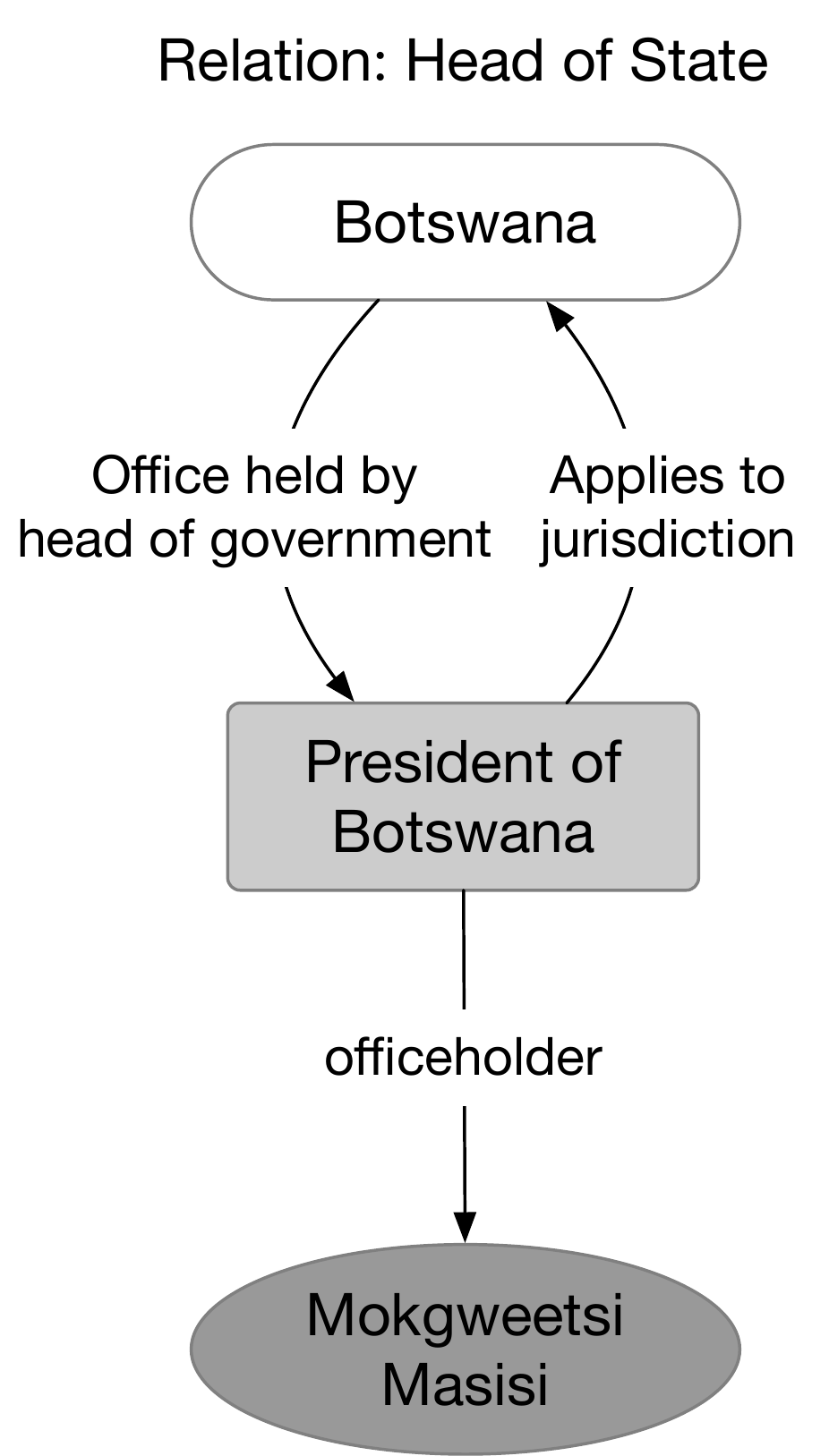}
		\caption{Path}
		\label{subfig:path}
	\end{subfigure}
	\begin{subfigure}[t]{0.35\linewidth}		
	\centering
	\includegraphics[width=\linewidth]{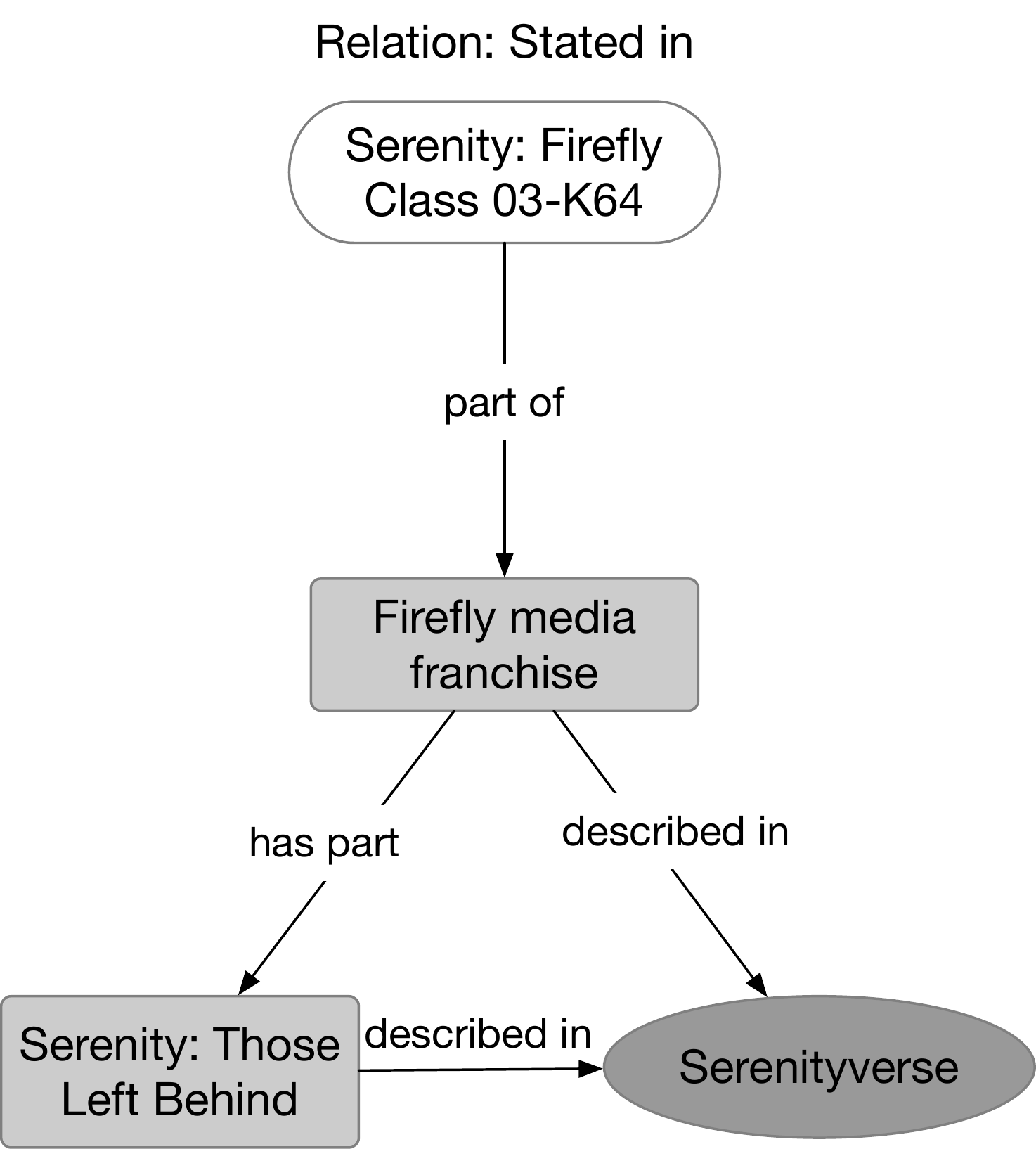}
	\caption{Triangle}
	\label{subfig:triangle}
		\end{subfigure}
	\begin{subfigure}[t]{0.3\linewidth}
		\centering
		\includegraphics[width=\linewidth]{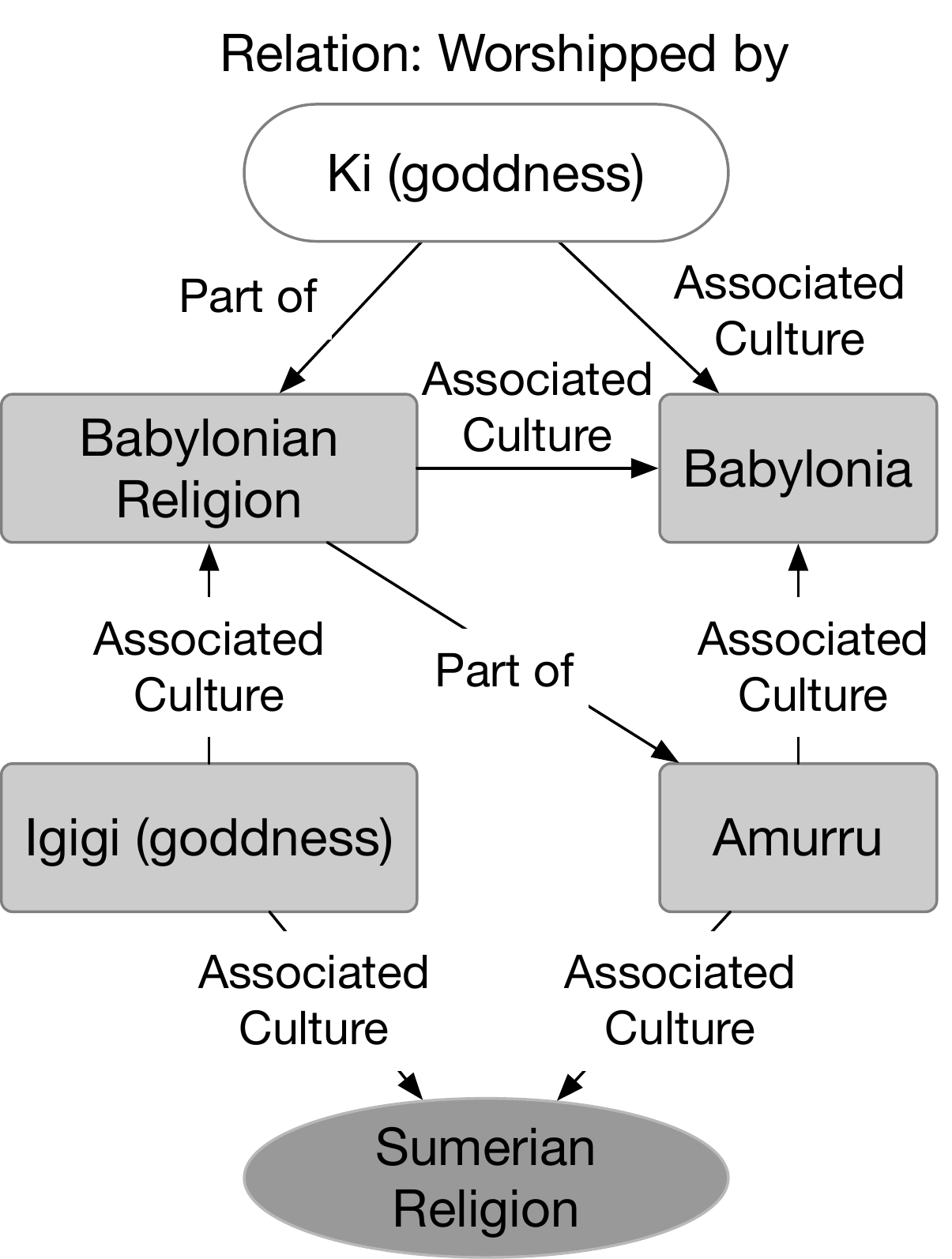}
		\caption{Graph}
		\label{subfig:graph}
	\end{subfigure}
	\caption{Case Study: different forms of reasoning graphs in the experiments. Capsules denote query entities, eclipses denote final answers, and rectangles denote intermediate nodes.}
	\label{fig:case}
\end{figure}

We show how \model\space can provide explainable reasoning graphs in the experiments in \Cref{fig:case}. These reasoning graphs are generated by selecting the subgraphs between query entities and final answers in cognitive graphs. The reasoning graph in \Cref{subfig:path} is a path from the head entity to the tail entity. However, the circle between "Botswana" and "President of Botswana" strengthens the reasoning process. \Cref{subfig:triangle} illustrates that multiple paths can boost the robustness of the answer. \Cref{subfig:graph} is an elaborate reasoning graph that contains various paths, triangles, and circles, which cannot be entirely modeled by path-based methods. \model\space utilizes the interacted information in such complex graphs to predict correct answers with more confidence.


%
%
\section{Related Work}
\label{sec:related}
\textbf{Knowledge Graph Embedding}\hspace{2mm} Embedding methods represent entities as continuous vectors, and various score functions are defined for a tuple $(\headentity, \relation, \tailentity)$, such as vector difference \cite{TransE,TransR:LinLSLZ15}, vector product \cite{DistMult:Yang2015,Complex:TrouillonDGWRB17}, convolution \cite{ConvE:DettmersMS018, Hypernet:abs-1808-07018}, and tensor operation \cite{TuckER, Rotate:sun2018}. Although these embedding approaches have achieved impressive results on several KG completion benchmarks, they have been shown to suffer from cascading errors when modeling multi-hop relations \cite{VariationalReasoning:ChenXYW18, Traverse:GuML15}, which are indispensable for more complex reasoning tasks.	Also, these methods all operate on latent space, and their predictions are not interpretable.

\textbf{Knowledge Graph Reasoning}\hspace{2mm}  Many works \cite{DeepPath:XiongHW17, GoForAWalk, M-Walk,VariationalReasoning:ChenXYW18} have proposed approaches that explicitly model multi-step paths for KG reasoning. The Path-Ranking Algorithm (PRA) uses a random walk with restart mechanism to obtain paths. Chain-of-Reasoning \cite{ChainOfReasoning:McCallumNDB17} and Compositional Reasoning \cite{CompositionalReasoning:NeelakantanRM15} take multi-hop paths found by PRA as input and aim to infer its relation. Two recent works, DeepPath \cite{DeepPath:XiongHW17} and MINERVA \cite{GoForAWalk}, use RL-based approaches to explore KG and find better reasoning paths. Later works extend MINERVA with reward reshaping \cite{MultihopKG:LinSX18} or Monte Carlo Tree Search \cite{M-Walk} respectively. \citet{VariationalReasoning:ChenXYW18} propose to unify path-finding and path-reasoning with variational inference. Our proposed model bases reasoning on subgraphs rather than paths, which can better capture the interaction of different paths. Another line of work is IRN \cite{IRN:ShenHCG17} and NeuralLP \cite{NeuralLP:YangYC17}, which learn first-order logical rules for KG reasoning with neural controller systems with external memory. However, these models often contain computationally expensive operations such as accessing the entire KG.

\textbf{Few-shot Learning}\hspace{2mm}
One-shot learning \cite{MatchingNetwork:Vinyals,ProtoNet:Snell,MAML:Finn,Meta-LSTM:Ravi} aims to complete the learning tasks with only a few training instances, such as image classification\cite{OneShot:Fei-FeiFP06} or machine translation\cite{MetaNMT:GuWCLC18}, which often require a large amount of data for traditional algorithms. This is often achieved by learning on a range of tasks. Previous work on one-shot learning can be divided into two groups: the metric methods \cite{SiameseNetwork:koch2015, MatchingNetwork:Vinyals, ProtoNet:Snell} that learn a similarity metric between new instances and instances in the training set, and the parameter methods \cite{MAML:Finn, Meta-LSTM:Ravi, MetaNetwork:MunkhdalaiY17} that directly predict or update parameters of the model according to the training data. Recently one-shot learning has been successfully applied in KG completion \cite{Oneshot-Relational:XiongYCGW18}, by learning a similarity metric with a single-layer graph convolutional network \cite{GCN:KipfW17}. However, their method still belongs to embedding-based methods and lacks multi-hop reasoning and interpretability. Also, their model requires forward pass through the neural network for every candidate, which is computationally expensive or even intractable for large-scale KGs.
\section{Conclusion}
\label{sec:conclusion}

We present a new framework \model\space to tackle one-shot KG reasoning problem at scale. The one-shot relational learning problem is solved with the combination of two modules, the summary module to summarize the underlying relationship of the given support pair and the reasoning module to find the correct answer based on the summary.
Under the inspiration of the dual process theory in cognitive science, we organize the reasoning process with a cognitive graph, achieving more powerful reasoning ability than previous path-based methods. Experimental results demonstrate the superiority of our framework. We also find that our method suffers from the non-connectivity of KGs. Therefore, in future work, we intend to improve the System 1 by allowing expanding unconnected nodes.

\bibliographystyle{plainnat}
\bibliography{reference}

\clearpage
\begin{appendices}
\section{Algorithm Implementation Details}

For the proposed \model, we set the dimensions of both entity embeddings and relation embeddings to 100 on NELL-One and 50 on Wiki-One. The dimension of hidden representations is set to 100 in both datasets. The maximum degree limit $\maxneighbor$ is set to 256 and the maximum node number $\lambda$ is set to 128. The action budget $\topk$ is set to 5. We use the ADAM optimization algorithm for model training with learning rate 0.00001 for entity and relation embeddings and 0.0001 for all the other parameters. We also add $L2$ regularization with weight decay 0.0001. The batch size is 32 on both datasets. We use the MRR on validation set as the standard for early-stop policy. On both datasets, we use pretrained embeddings generated by DistMult \cite{DistMult:Yang2015}.

The model is implemented with PyTorch 1.1\footnote{\url{https://pytorch.org/}}. The source code is also provided in the supplementary material. We run all the experiments on a single Linux server with 8 NVIDIA RTX 2080. 

\section{Experiment Details}	
\subsection{Datasets}

\begin{table}[!h]
	\caption{Statistics of datasets used in experiments.}
	\label{tab:dataset}
	\centering
	\begin{tabular}{lrrrr}
		\toprule
		Dataset & \#entities & \#relations & \# Triples & \# Tasks (Train/Valid/Test)\\
		\midrule
		NELL-One & 68,545 & 358 & 181,109 & 67 (51/5/11)\\
		Wiki-One & 4,838,244 & 822 & 5,859,240 & 183 (133/16/34)\\
		\bottomrule
	\end{tabular}
\end{table} 

We use the NELL-One and Wiki-One datasets released by \citet{Oneshot-Relational:XiongYCGW18} for evaluation\footnote{\url{https://github.com/xwhan/One-shot-Relational-Learning}}. The dataset statistics are shown in \Cref{tab:dataset}. Both datasets are created with a similar process: relations with less than 500 but more than 50 triples are selected as one-shot tasks and randomly divided into training, validation, and testing relations. The background KGs are built with facts of other relations.

In practice, we found that the Wiki-One dataset suffers from sparsity and non-connectivity in the backend KG. In the test set, 15.8\% of the entity pairs are not connected at all and the distances of other 25.5\% pairs are no less than 5. For these 41.3\% pairs, we don not have any reasonable paths to infer their relations. To better evaluate the model's reasoning ability, we remove the evaluation facts whose entity pairs' distances are equal to or more than 5 in Wiki-One.

\subsection{Baselines}
\paragraph{Implementation} For TransE, ComplEx and DistMult, we use the implementation\footnote{\url{https://github.com/DeepGraphLearning/KnowledgeGraphEmbedding}} released by \citet{Rotate:sun2018}. For ConvE and MultihopKG, we use the implementation\footnote{\url{https://github.com/salesforce/MultiHopKG}} released by \citet{MultihopKG:LinSX18}. For TuckER \cite{TuckER}, we use the implementation released by the author\footnote{\url{https://github.com/ibalazevic/TuckER}}. For GMatching, we use the implementation and pretrained embeddings released by the author\footnote{\url{https://github.com/xwhan/One-shot-Relational-Learning}}. 

\paragraph{Hyperparameter} For all the embedding-based methods, the embedding dim is set to 100 on NELL-One and 50 on Wiki-One, which is consistent with the settings of our method and GMatching. For MultihopKG, we use pretrained embeddings generated by ConvE, which achieves best results in their experiment.

\paragraph{Setting} For all the embedding-based methods and MultihopKG, we use the triples of background relations, all the triples of the training relations, and the ont-shot training triples of validation/test relations for training. For GMatching, we follow the one-shot learning setting described in their paper. Note that unlike GMatching, our method does not need a separate set of training relations except the relations in the background KG. Therefore we simply merge the training relations into the background KG.

\paragraph{Performance} Considering the relatively small scale of NELL-One, we run each method three times and report the mean and stddev. On Wiki-One, we only run each method once since the scale of the dataset is quite large and the margins among different methods are quite significant. For TransE, ComplEx and DistMult on Wiki-One, our experiment gives much worse results than those reported in \citet{Oneshot-Relational:XiongYCGW18}, so we quote their results in the paper.

\section{Additional Experiments}
\begin{table}
	\caption{Results on Wiki-One(not filtered).}
	\label{tab:wikidata}
	\centering
	\begin{tabular}{lrrrr}
		\toprule
		Model & Hits@1 & Hits@5 & Hits@10 & MRR\\
		\midrule
		GMatching(Best) & 12.0 & \textbf{27.1} & \textbf{33.6} & \textbf{20.0}\\
		\model & \textbf{14.6} & 19.4 & 21.3 & 16.8\\
		\bottomrule
	\end{tabular}
\end{table}
We provide the experimental results against GMatching in the original Wiki-One dataset in \Cref{tab:wikidata}. We observe that on this dataset our model achieves the highest Hits@1 score while shows weaker performance compared to GMatching in terms of Hits@5, 10 and MRR.

The difference comes from the unreachable cases filtered in our dataset, which are unfriendly for path-based solutions. This will also harm the performance of most embedding-based methods. The GMatching method, however, is less influenced by the non-connectivity because the graph convolutional network for query pair matching can reduce the candidate space by entities' local patterns (e.x, shared relation edges in training pairs and test pairs). This can also explain why GMatching can beat our model on Hits@10 but not on Hits@1. The ability to predict the true tail entities that are not directly connected with the head entity, although possibly crucial for KG completion, is beyond the scope of the paper. From the point of logic, there are not sufficient evidences to reason the relationship of entities that are not connected in the KG. If we do care about such entity pairs, the remedy, however, is simple that we can ensemble an embedding based method to solve the unreachable cases.

\section{Further Discussion}
\subsection{Time Complexity}
In \citet{Oneshot-Relational:XiongYCGW18}, they argue that the candidate set for a query relation can be constructed using the entity type constraint, which makes their relatively complex matching model feasible for large datasets like Wiki-One. However, from their released code we can find two obvious difficulties for doing so. Firstly, it's not always possible to construct the entity type constraint. For some datasets, like FB15k-237 \cite{FB15k:ToutanovaCPPCG15}, the information of entity types is missing. And to cover all the possible entity types for a relation, we have to enumerate all the facts, including the evaluation ones. Secondly, even if we build such constraint, the candidate set can still be very large, or even equal to the entity set. For example, on Wiki-One, they have to truncate the candidate sets to 5000 for some relations. Therefore, the time complexity is significant for applying KG completion algorithms on large-scale KGs.

\end{appendices}

\end{document}